# Learning Implicit Brain MRI Manifolds with Deep Learning

Camilo Bermudez*[a], Andrew J. Plassard[b], Larry T. Davis[c], Allen T. Newton[c],
Susan M Resnick[b], Bennett A. Landman[a]

[a] Department of Biomedical Engineering, Vanderbilt University, 2201 West End Ave, Nashville, TN, USA 37235;
[b] Department of Computer Science, Vanderbilt University, 2201 West End Ave, Nashville, TN, USA 37235;
[c] Department of Radiology, Vanderbilt University Medical Center, 2201 West End Ave, Nashville, TN, USA 37235;
[d] Laboratory of Behavioral Neuroscience, National Institute on Aging, Baltimore, MD, USA

## ABSTRACT

An important task in image processing and neuroimaging is to extract quantitative information from the acquired images in order to make observations about the presence of disease or markers of development in populations. Having a low-dimensional manifold of an image allows for easier statistical comparisons between groups and the synthesis of group representatives. Previous studies have sought to identify the best mapping of brain MRI to a low-dimensional manifold, but have been limited by assumptions of explicit similarity measures. In this work, we use deep learning techniques to investigate implicit manifolds of normal brains and generate new, high-quality images. We explore implicit manifolds by addressing the problems of image synthesis and image denoising as important tools in manifold learning. First, we propose the unsupervised synthesis of T1-weighted brain MRI using a Generative Adversarial Network (GAN) by learning from 528 examples of 2D axial slices of brain MRI. Synthesized images were first shown to be unique by performing a cross-correlation with the training set. Real and synthesized images were then assessed in a blinded manner by two imaging experts providing an image quality score of 1-5. The quality score of the synthetic image showed substantial overlap with that of the real images. Moreover, we use an autoencoder with skip connections for image denoising, showing that the proposed method results in higher PSNR than FSL SUSAN after denoising. This work shows the power of artificial networks to synthesize realistic imaging data, which can be used to improve image processing techniques and provide a quantitative framework to structural changes in the brain.

## 1. INTRODUCTION

An important task in image processing and neuroimaging is to extract quantitative information from the acquired images in order to make observations about the presence of disease or markers of development in populations. Having a low-dimensional representation of an image allows for easier statistical comparisons between groups. Gerber et. al showed that a low-dimensional, non-linear manifold can effectively represent the variability found in brain anatomy [1]. The mapping of a brain image into a manifold and vice versa can help construct priors that span the entire variability spectrum of a normal human brain. Many groups have sought to identify the best brain representation, such as building a template or atlas [2], linear mappings [3], Isomap [1], or regressions based on underlying parameters such as size or shape of regions of interest [4]. However, these approaches are limited, first in that the volumetric ROI measures do not provide a comprehensive mapping to the subject space. Second, a linear representation does not account for nonlinear effects present in the human brain such as aging [5]. Lastly, a generative nonlinear mapping such as Isomap require explicit similarity measures and can only generate images from the manifold by interpolating from examples, resulting in overly smoothed images. We posit that the ideal manifold mapping would learn a naive representation using the entire image and from this manifold generate images that are of comparable quality as the learning set.

Recently, deep learning approaches like deep autoencoders and convolutional neural networks have provided a new mechanism for learning a manifold representation of data [6, 7]. These approaches calculate non-linear functions between inputs and can either be supervised, in the case of approaches like deep convolutional networks, or unsupervised, in the case of autoencoders [8]. Deep learning approaches are generally based on neural networks, where there are a series of layers either sparsely or densely connected between them. In this work, we propose a method of implicit manifold learning of brain MRI through two common image processing tasks: image synthesis and image denoising.

Past research on image synthesis has focused on two problems: increasing image resolution and inter-modality image generation. Image super-resolution seeks to learn the map from low-resolution images to high resolution images [9]. Similarly, inter-modality image synthesis seeks to generate one modality from another, as in the case of CT from MRI, to avoid excess radiation or impractical sequences [10-12]. Nevertheless, image synthesis is a form of manifold learning from one image space to another. Previous work in this field has resulted in images that look near-realistic, but have not been validated for image quality by imaging experts. The second use case for manifold learning is image denoising. This tool is ubiquitously used in image processing to improve the performance of registration and segmentation algorithms by increasing the signal-to-noise ratio. In manifold learning, image denoising allows for a better mapping from image space to the manifold. Therefore, the distance metric in manifold space can better discriminate differences between brain representations.

In this work, we use deep learning techniques to explore the manifold of normal brains and generate new, high-quality images. We explore the implicit manifold by addressing the problems of image synthesis and image denoising as important tools in manifold learning. To do so, we use propose the use of Generative Adversarial Networks (GANs), which have shown to produce high-resolution, high-fidelity images in an unsupervised manner in the field of computer vision [8, 13]. By training two coupled networks, a generator network, which synthesizes images that resemble the training set, and a discriminator network, which learns to classify an image as real or fake, we are able to produce high-fidelity images that closely resemble acquired images. We also propose the use of skip-connected autoencoders for image denoising, which have been shown to perform well in this task in the field of computer vision [14]. The connections between convolutional layers in the autoencoder preserve structural features to increase resolution. We show that this denoising technique outperforms current state-of-the art denoising software FSL SUSAN. Together, these two tools help develop a mapping to an implicit brain MRI manifold in order to generate realistic MRI images that span the domain of a normal brain.

## 2. METHODS

### 2.1 Dataset

The image synthesis GAN and the denoising autoencoder experiments were conducted on 528 T1-weighted brain MRI images from healthy controls (ages 29-94 years old, mean of 67.9 years old) as part of the Baltimore Longitudinal Study of Aging (BLSA) study, which is a study of aging operated by the National Institute of Aging [15]. MR images were acquired on a Phillips 3T scanner using an MPRAGE sequence and with 1mm isotropic voxel resolution. All subjects were affine-registered to MNIs-space and intensity-normalized. For computation efficiency, the single midline axial slice was chosen for each subject to train the model. The dimensions of all slices were 220 x 172 voxels.

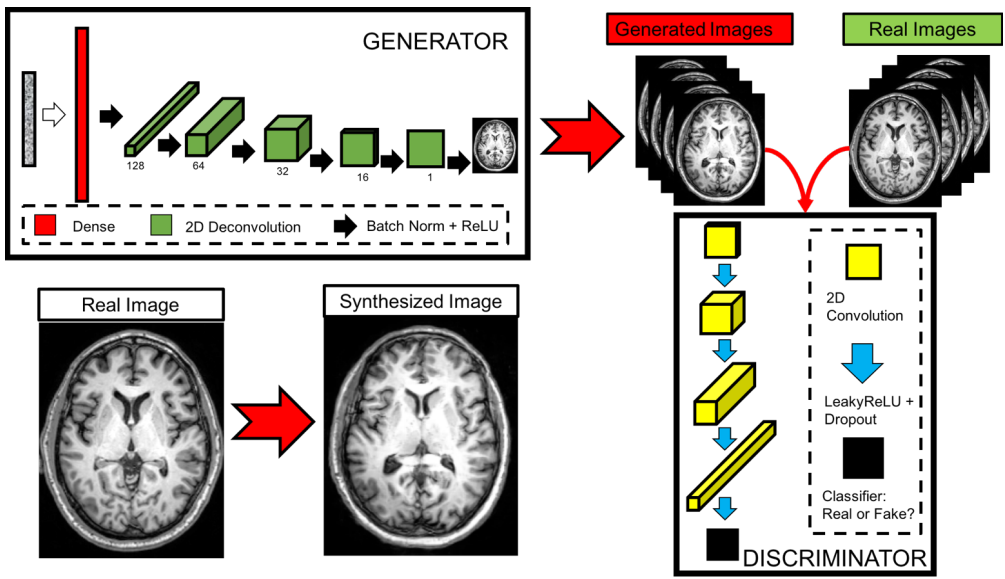

Figure 1. Pipeline and network architecture for Generative Adversarial Network used for image synthesis. The generator is a 2D deconvolutional neural network that takes noise as input and generates a 2D image of the brain. The discriminator is a convolutional neural network that takes real and synthetic images and learns to classify them.

## 2.2 Generative Adversarial Networks for Image Synthesis

The architecture of the generator and discriminator used in our GAN model is based on [13]. The input for the generator is a one-dimensional vector of uniform noise with values between -1 and 1. The first step is a dense layer that reshapes increases the dimensionality of the input vector. Then, we use a series of upsampling, 2D deconvolution of increasing filters, and batch normalization layers to restore the image to the size of the training set (Figure 1). Each deconvolutional layer consists of a kernel size of 3 pixels and a stride of 1 pixel. All layers used ReLU activation functions.

The discriminator takes as inputs both image sets: the real images and the synthesized images, as well as the corresponding labels. Instead of generating an image, the discriminator learns an abstract representation following an inverse network architecture: 4 2D-convolutional layers of increasing filter size along with LeakyReLU and Dropout layers after each convolution (Figure 1). The convolution has a kernel size of 3 pixels and a stride of two pixels to decrease the dimensionality. The purpose of the Dropout and LeakyReLU layers is to avoid overfitting to the training set [13]. The last layer is a dense layer with one node and a sigmoid activation function, which acts as a binary classifier for each input. By coupling the generator and the discriminator, the GAN will maximize the classification accuracy of true image versus synthetic image, while also forcing the generator the produce more realistic images with each epoch.

This network was trained for 1500 epochs using the Adam optimizer with a learning rate of 0.0001. The training set consisted of 528 noise vectors, while the testing set consisted of 100 new noise vectors. An image correlation was performed between the synthesized images and the testing set to ensure that the network was not learning the null manifold and synthesizing examples from training. Next, we validated the quality of synthesized images with two raters, a neuroradiologist and an imaging expert, who scored image quality 1-5 on a random deck of 100 true and 100 synthesized images.

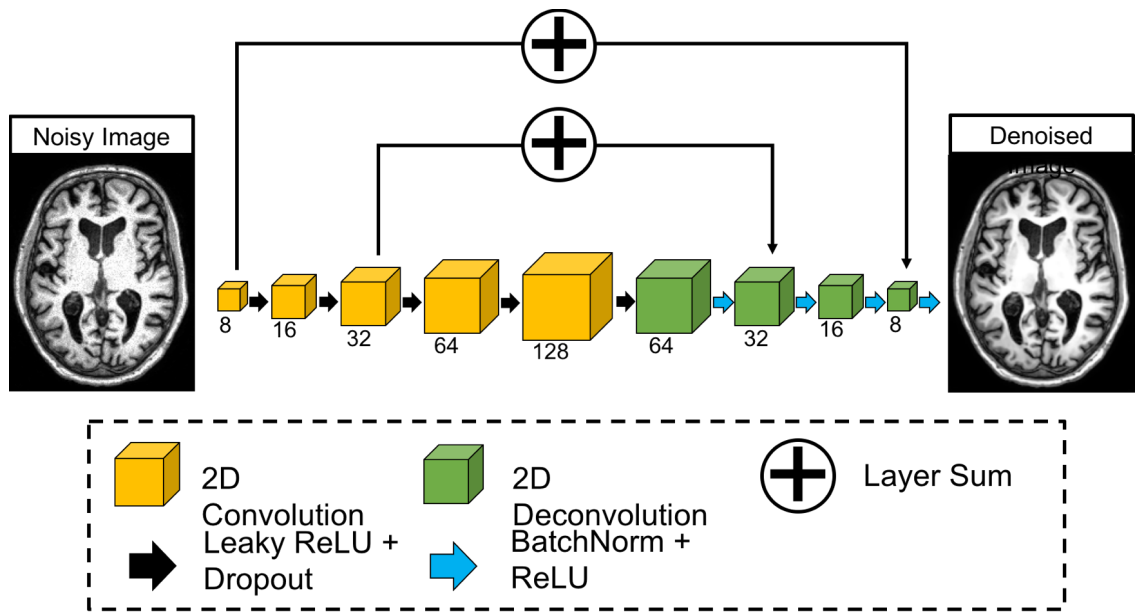

Figure 2. Denoising Autoencoder Network. This network consists of a convolutional neural network of increasing filter size, followed by a deconvolutional neural network of decreasing filter size. It takes a noisy image as the input and returns the denoised image.

## 2.3 Denoising Autoencoder

First, noisy images were generated by adding Gaussian noise within a mask of the brain and skull to the 2D slices of our dataset. Three different noisy images were generated for each subject with SNR of 1, 10, and 100. Three separate instances of the proposed autoencoder network were trained for each level of noise. The proposed network takes as input a noisy image and encodes it into a high-dimensional representation through a series of 2D-convolutions with increasing filters (Figure 2). This representation is then returned to the original image using 2D-deconvolution. The goal of this network is to minimize the mean squared error between the denoised image and the original image. In order to avoid smoothing and preserve the structural features, we add the feature output from two convolutional layers to its corresponding deconvolutional layers as proposed by [14].

We validate this denoising technique by comparing it to FSL SUSAN v5.0. First, we perform a grid search to find the optimal parameters of brightness threshold and kernel size in SUSAN working on the noisy images with SNR of 10. Once optimized, these parameters were used to denoise the noisy images with SNR of 1 and 100. PSNR was calculated for the input noisy images, SUSAN images, and the proposed method.

## 3. RESULTS

### 3.1 Brain MRI Synthesis

The proposed network is able to generate realistic brain images that closely resemble the training set. Figure 3 shows a correlation matrix between training images and the images synthesized from training noise. A representative synthesized image, as well as three real images with highest correlation values are shown in Figure 3. The real images with high correlation show several structural differences from the synthesized images, such as ventricle size or gyral patterns, suggesting that this network is not simply reconstructing examples from the training set.

Validation of the synthesized images by two separate raters shows an average quality score of 4.9 for the real image and a quality score of 3.9. However, Figure 4 shows the histogram distribution of scores between real and synthetic images by each rater. In the case of both raters, there is substantial overlap in the distribution of scores, resulting in many synthetic images given scores 3 or higher and real images with a score of 2 or lower.

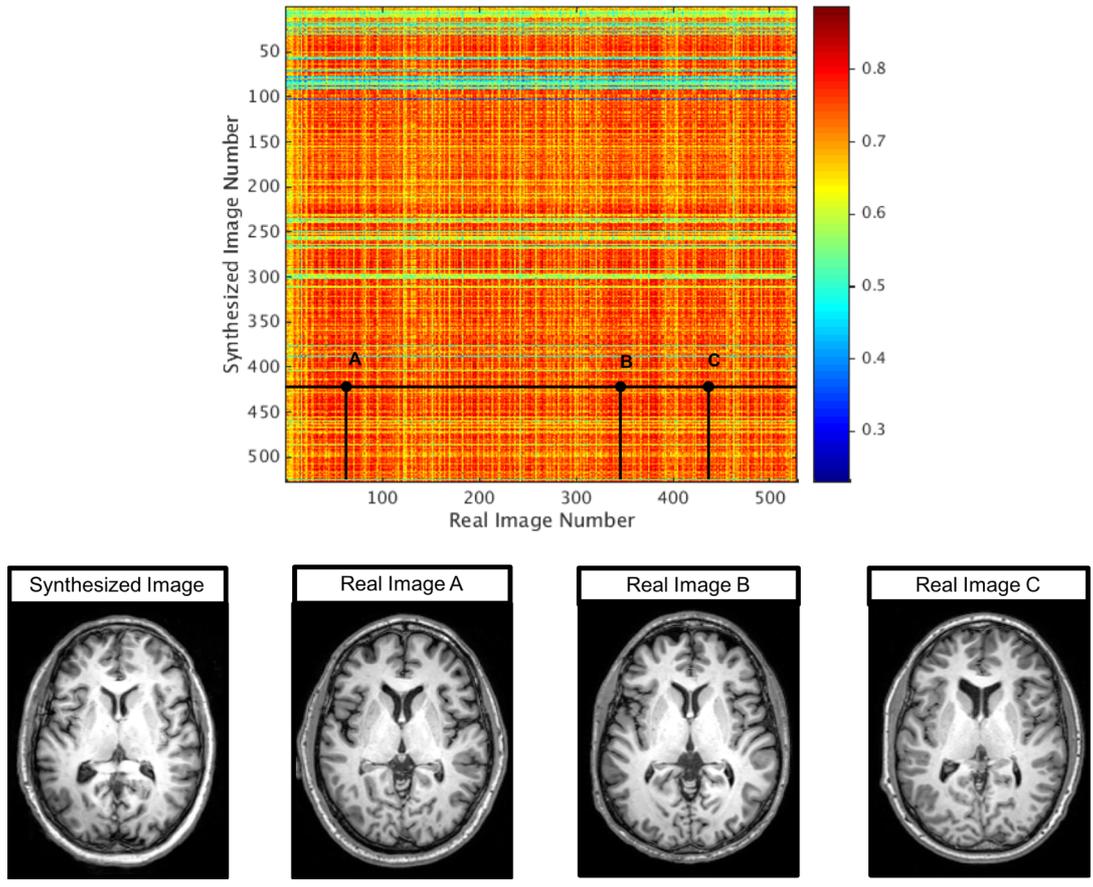

Figure 3. Correlation map between synthesized images and real images. Representative examples a synthetic image (top left) and three real images with the highest correlation values: Real Image A (rho = 0.76), Real Image B (rho = 0.77), and Real Image C (rho = 0.78).

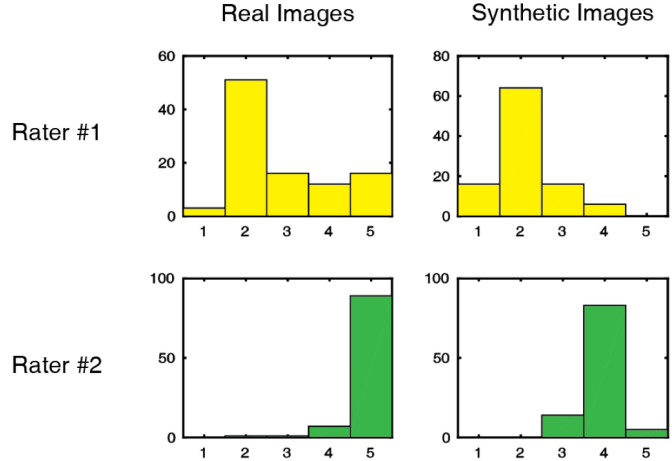

Figure 4. Histograms of image quality score provided by imaging expert. Rater #1 (above) is a neuroradiologist and Rater #2 (below) is an imaging expert.

### 3.2 Image Denoising

Our proposed method is able to denoise images better than current state-of-the-art algorithm FSL SUSAN. Figure 5 shows that high-fidelity image reconstruction is achieved when the SNR is over 10. To validate the results, PSNR was measured in the denoised image from our method as well as with FSL SUSAN. The proposed method showed a greater PSNR compared to FSL SUSAN using the Wilcox rank sum test ($p < 0.001$) for all three levels of noise.

## 4. DISCUSSION & CONCLUSION

We show that the proposed method is able to generate images of brain MRI that closely resemble those in the training set. We showed that our method does not simply replicate the null set by learning an example from the training images. Instead, this technique generates new brain images that belong to the manifold of brain MRI slices. Moreover, we were able to show that the generated images can be rated with comparable quality as real images. Of note, the first rater, a neuroradiologist, mentioned that despite comparable quality, the synthetic images were immediately given away by anatomic abnormalities such as largely asymmetric left and right caudate. Similarly, the second rater, a neuroimaging expert, noticed brighter intensities near the center of the image compared to the boundaries in the synthetic images. This would immediately be noticed after acquisition by the technician and reported as a hardware issue. These comments represent challenges in image synthesis: anatomic accuracy and signal quality.

In the case of manifold learning, anatomic accuracy could be attained by imposing structural restrictions, such as the inter-layer connections used in the autoencoder or symmetry connections within layers. This leads to a more refined notion of the normal brain manifold. If our method can produce realistic images that are not anatomically viable, then further work is needed to refine the subset that truly represents possible MRI images. However, exploration of these unrealistic synthesized images may shed a light on possible structural and functional variants in brain anatomy found in healthy individuals or disease. Secondly, the problem of signal quality is essential to manifold learning, since it allows for better distinction between subjects. Physical restrictions can again be applied to make the image intensity homogenous and realistic. Alternatively, this might represent artifact found in the training data that is not observed in single images, but highlighted by its synthesized representatives. In the task of image denoising we show that our method is superior to FSL SUSAN across all SNR levels. Image fidelity is recovered at higher SNR, but smoothing effects are seen in SNR of 1. Further work would require denoising 3D brain volumes, but this work may encounter resource challenges due to the depth of the network and the high dimensionality of the problem.

It is remarkable that the two uses of image synthesis and image denoising achieved promising results while using a dataset of only 528 images. This is probably due to having a relatively homogeneous training set in terms of acquisition parameters and demographics, aimed at solving a simple problem. However, the deep learning field has shown that results improve remarkably with over 10,000 training examples [8]. The work presented here would show better results if the respective networks were trained with 3D brain volumes from subjects of all ages, from different centers, and with multiple acquisition parameters. In fact, including this information as contextual features in the training process would further increase image fidelity. Using these tools, the construction of a comprehensive normal brain manifold would allow for quantitative exploration of structural and functional diseases that can be easily implemented into clinical practice.

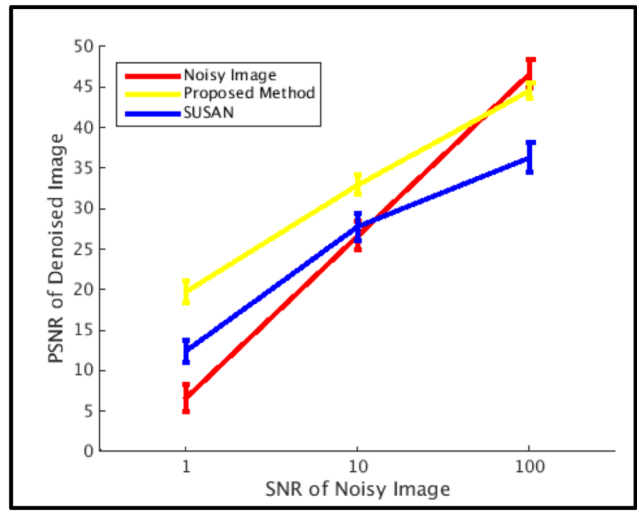

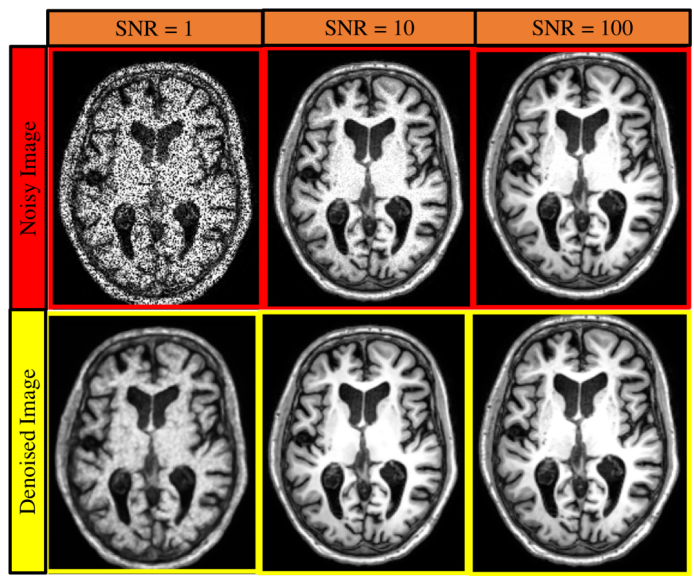

Figure 5. Image denoising results. The left column shows representatives of noisy input images at different SNR levels of 1, 10, and 100. The right column shows the denoised result with the proposed method. The graph shows PSNR of the input image, the output of FSL SUSAN, and the output of the proposed method. Our method shows higher reconstructed PSNR across all levels of noise ($p < 0.01$).

## ACKNOWLEDGEMENTS

This research was supported by NSF CAREER 1452485, NIH grants 1R03EB012461 (Landman), R01NS095291 (Dawant), U54 HD083211 (Dykens; NIH/NICHD) 5R01 HD044073 (Cutting; NIH/NICHD), 5R01 HD067254 (Cutting; NIH/NICHD), and T32GM007347 (NIGMS/NIH). This research was conducted with the support from Intramural Research Program, National Institute on Aging, NIH. We are appreciative of the volunteers who gave their time and de-identified imaging data to be a part of this study. This study was in part using the resources of the Advanced Computing Center for Research and Education (ACCRE) at Vanderbilt University, Nashville, TN. This project was supported in part by ViSE/VICTR VR3029 and the National Center for Research Resources, Grant UL1 RR024975-01, and is now at the National Center for Advancing Translational Sciences, Grant 2 UL1 TR000445-06. This work does not reflect the opinions of the NIH or the NSF.